\documentclass[runningheads]{llncs}

\usepackage{eccv}

\usepackage{eccvabbrv}

\usepackage{graphicx}
\usepackage{booktabs}
\usepackage{algorithm}
\usepackage{algpseudocode}
\usepackage{amsmath}
\usepackage{bm}
\usepackage{xcolor}

\newcommand{\boldcheckmark}{%
  \textcolor{black}{\checkmark}%
  \kern-1.4ex\textcolor{black}{\checkmark}%
}
\usepackage[accsupp]{axessibility}  % Improves PDF readability for those with disabilities.

\usepackage[pagebackref,breaklinks,colorlinks,citecolor=eccvblue]{hyperref}
\usepackage{hyperref}

\usepackage{orcidlink}

\begin{document}

\title{AnatoMask: Enhancing Medical Image Segmentation with Reconstruction-guided Self-masking} 

\titlerunning{AnatoMask for medical image segmentation}

\author{Yuheng Li\inst{1,2}\orcidlink{0009-0001-0249-8952} \and
Tianyu Luan\inst{3} \and
Yizhou Wu\inst{2} \and
Shaoyan Pan\inst{1} \orcidlink{0009-0007-1040-0189}\and
Yenho Chen\inst{2} \and
Xiaofeng Yang\inst{1,2} \orcidlink{0000-0001-9023-5855}}
\authorrunning{Li et al.}

\institute{Emory University, Atlanta GA 30322, USA \and
Georgia Institute of Technology, Atlanta GA 30332, USA \and
State University of New York at Buffalo, Buffalo, NY, 14260 \\ 
\email{xiaofeng.yang@emory.edu}}

\maketitle

\begin{abstract}
Due to the scarcity of labeled data, self-supervised learning (SSL) has gained much attention in 3D medical image segmentation, by extracting semantic representations from unlabeled data. Among SSL strategies, Masked image modeling (MIM) has shown effectiveness by reconstructing randomly masked images to learn detailed representations. However, conventional MIM methods require extensive training data to achieve good performance, which still poses a challenge for medical imaging. Since random masking uniformly samples all regions within medical images, it may overlook crucial anatomical regions and thus degrade the pretraining efficiency. We propose AnatoMask, a novel MIM method that leverages reconstruction loss to dynamically identify and mask out anatomically significant regions to improve pretraining efficacy. AnatoMask takes a self-distillation approach, where the model learns both how to find more significant regions to mask and how to reconstruct these masked regions. To avoid suboptimal learning, Anatomask adjusts the pretraining difficulty progressively using a masking dynamics function. We have evaluated our method on 4 public datasets with multiple imaging modalities (CT, MRI, and PET). AnatoMask demonstrates superior performance and scalability compared to existing SSL methods. The code is available at \href{https://github.com/ricklisz/AnatoMask}{https://github.com/ricklisz/AnatoMask}.

\end{abstract}

\section{Introduction}
\label{sec:intro}
While supervised learning shows great promise in 3D medical image segmentation, its potential in clinical settings is hindered by a lack of labeled data. As an alternative, self-supervised learning has shown great promise in medical image analysis by extracting semantic representations from unlabeled data \cite{tang2022self,  jiang2023anatomical, zhou2021preservational,zhu2020rubik}. Among various SSL methods, masked image modeling has achieved promising results in natural images \cite{he2022masked} and has been applied in medical image analysis \cite{chen2023masked, zhou2023self, chen2022multi}. MIM utilizes a simple task in which the model is asked to reconstruct parts of the masked image, thereby learning detailed representations of the image \cite{xie2022simmim, he2022masked}.

However, MIM's success in natural images is heavily dependent on large-scale training data \cite{xie2023data}, which poses a significant challenge in medical imaging. While many MIM methods utilize random masking strategy and achieved success in natural images \cite{he2022masked, bao2021beit}, we argue that this strategy does not utilize the inherent properties of anatomical regions within medical images and leads to less effective pretraining. In medical images, anatomical regions such as organs and tumors, are deemed more important than the less relevant air-filled and fluid-filled regions. Since anatomical regions are characterized by varying shapes and contrasts, these regions are intuitively more difficult to reconstruct when masked out, requiring more guidance during pretraining. 
% Random masking, on the other hand, uniformly samples all regions and fails to account for the importance of anatomical regions, therefore requiring large-scale data to achieve good performance.
Thus, the previous methods require more training data to achieve the expected results, which greatly limits their application in medical-related tasks. Therefore, to design an effective MIM framework for medical images, it is crucial to identify these complex anatomical regions and create a more challenging task. 

\begin{figure}[tb]
  \centering
  \includegraphics[width=1\linewidth]{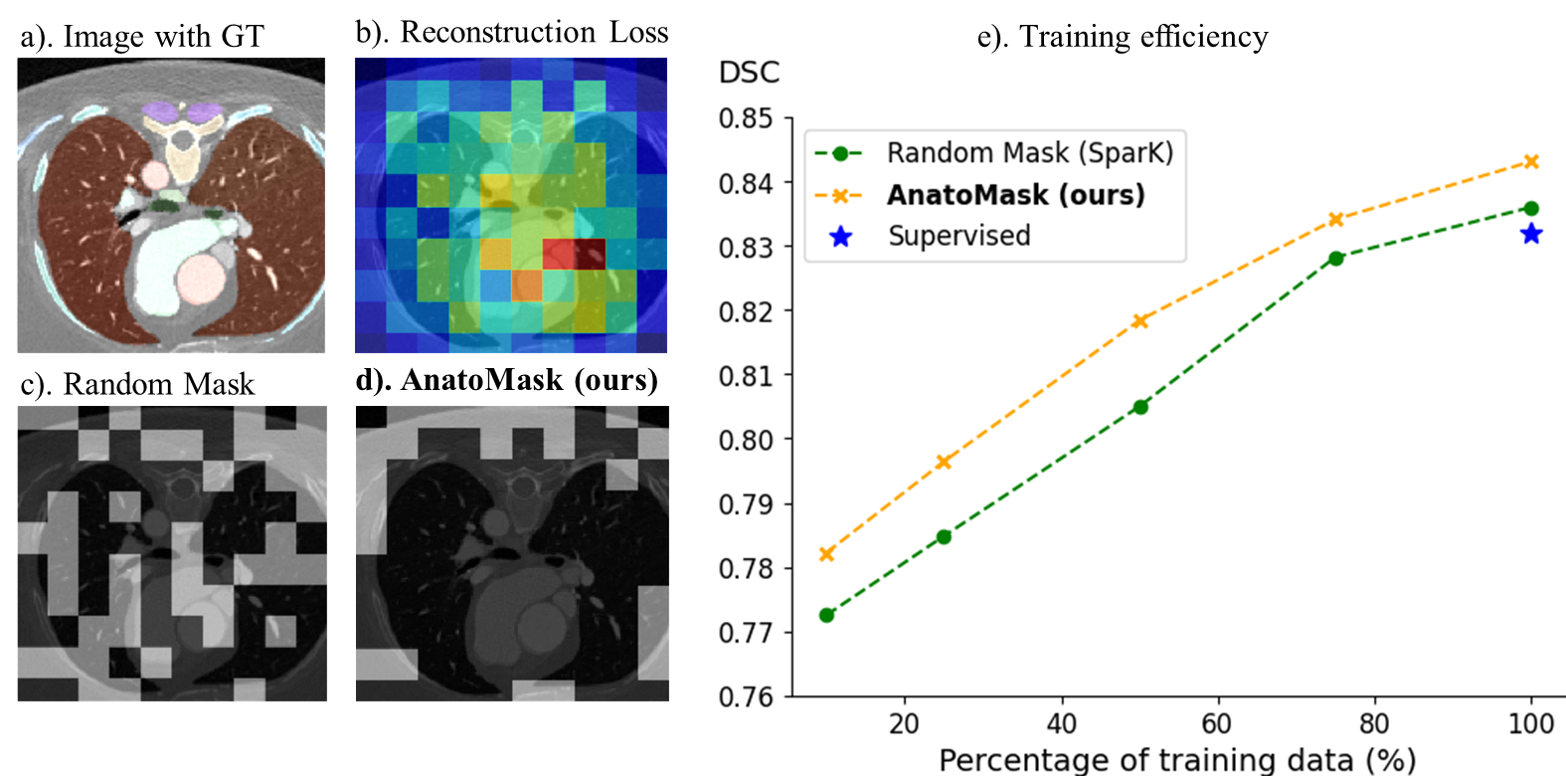}
  \caption{Comparison of random masking (SparK) vs AnatoMask. We visualize: a). input with anatomical ground truth (GT); b). reconstruction losses obtained by averaging over 2 random masks; c). a random mask ; d). our AnatoMask generated from b. In e), we also compared the training efficiency of AnatoMask and SparK. Transparent areas indicate unmasked regions. }
  \label{fig:AnatoMask_Fig1}
\end{figure}

To address these challenges, we propose AnatoMask, a reconstruction-guided MIM framework where the model learns from more informative masks to enhance pretraining efficacy. Observing that anatomical regions often featured higher reconstruction losses (\cref{fig:AnatoMask_Fig1} a and b),
% we argue that masking out such regions would allow the model to extract anatomical information more efficiently for subsequent segmentation tasks. 
if we can increase the probability of masking these regions, it would also encourage the network to extract information from these regions more efficiently.
Hence, we propose to identify anatomically-significant regions by reconstruction losses. Instead of using a pretrained model to identify these regions, we propose a self-distillation approach where the model can first act as a teacher to generate anatomically-significant masks, and then act as a student to learn from these masks. To ensure that the teacher is adaptive to the student’s performance, we update the teacher model with the student’s exponential moving average (EMA) weights. In this way, we not only encourage the model to learn how to reconstruct these regions but also where to find such important regions. This dual-learning approach is crucial in developing a deeper anatomical understanding, especially in data-constrained scenarios. Finally, we propose an easy-to-hard masking strategy, where the teacher continuously increases the difficulty of the MIM task as the student progresses, avoiding the model overly focusing on difficult regions at first. We conduct self-supervised learning on TotalSegmentator \cite{wasserthal2023totalsegmentator}, a comprehensive CT dataset for volumetric medical image segmentation. We adopt a previous SOTA architecture STU-Net \cite{huang2023stu} and evaluate the pretraining efficacy by finetuning on TotalSegmentator \cite{wasserthal2023totalsegmentator}, FLARE22 \cite{ma2022fast}, AMOS22 \cite{ji2022amos} and AutoPETII \cite{gatidis2022whole}. We first compare the differences between random masking and our AnatoMask in (\cref{fig:AnatoMask_Fig1} c and d). We then compare the training efficiency of AnatoMask vs random masking (SparK \cite{tian2023designing}) in (\cref{fig:AnatoMask_Fig1} e). Our method consistently improves the training efficiency, outperforming random masking. We also compare our method with previous state-of-the-art (SOTA) SSL approaches. We find that our method demonstrates superior performance and scaling capabilities.

We summarize our contribution as follows:
1). We propose a reconstruction-guided masking strategy, in which the model learns the anatomically significant regions through reconstruction losses and dynamically adjusts the masking to formulate a more challenging and specific MIM objective, thereby improving pretraining efficacy.  
2). We propose to leverage self-distillation to perform self-masking. The teacher network first identifies important regions to mask and generates a more difficult mask for the student to solve. The student will then update the teacher with EMA weights. 
3). To prevent the network from converging to a suboptimal solution early during training, we propose a masking dynamics function controlling the difficulty of the MIM objective, facilitating an easy-to-hard strategy

\section{Related Works}

\textbf{Contrastive learning.} 
Self-supervised learning leverages unlabeled data to learn meaningful representations without explicit supervision, demonstrating significant advancements in pattern recognition \cite{doersch2015unsupervised,baevski2022data2vec,gidaris2018unsupervised,oord2018representation,wang2015unsupervised,wu2018unsupervised,zhang2016colorful,noroozi2016unsupervised}. In this paradigm, contrastive learning first gains much attention in computer vision \cite{hjelm2018learning, misra2020self, chen2020simple, he2020momentum, grill2020bootstrap,gutmann2010noise} and medical imaging \cite{chaitanya2020contrastive, Taleb2021multimodal,tang2022self, azizi2021big}. Contrastive learning extracts powerful semantic representation by minimizing the distance of similar image pairs while maximizing dissimilar ones. However, contrastive learning relies on tailored data augmentations to create such image pairs \cite{chen2020improved}. Inappropriate augmentations could distort important clinical semantics leading to misrepresentations in the learned features. Also, since segmentation tasks require precise pixel-level embeddings, contrastive methods could overlook the nuanced features critical for detecting small organs or anomalies.
\\
\textbf{Masked image modeling for ViTs, ConvNets.} Motivated by the success of masked language modeling in natural language processing, masked image modeling has attracted much attention in computer vision. MIM learns image representations by reconstructing their masked parts, a concept that resembles early techniques like context encoders \cite{pathak2016context}. Naturally, MIM is first applied to vision transformers due to its sequence-to-sequence modeling. Both He et al. \cite{he2022masked} and Xie et al. \cite{xie2022simmim} propose to mask random image patches and regress the RGB pixels. More recent advancements in MIM focus on new reconstruction targets \cite{bao2021beit, wei2022masked, peng2022beit,tao2023siamese,huang2023contrastive} or masking strategies \cite{kakogeorgiou2022hide, shi2022adversarial, li2022semmae, wang2023hard}. For ConvNets, recent works such as SparK \cite{tian2023designing} and ConvNeXt V2 \cite{woo2023convnext} propose to leverage sparse convolution to address the information leakage during pretraining. However, novel MIM strategies for medical image segmentation remain largely unexplored.\\
\textbf{SSL for medical image segmentation.} Previous works propose various pretext tasks to solve medical image segmentation with promising performance. Similar to the context encoder, image inpainting has been applied to various ConvNet backbones to generalize the anatomical contents of medical images \cite{chen2019self}. Taleb et al. explore a variety of pretext tasks such as rotation, solving Rubik’s cube, and predicting patch locations \cite{taleb20203d}. Tang et al. propose a combination of pretext tasks and contrastive learning for self-supervising transformer \cite{tang2022self}. Fatemeh et al. combine discriminative, restorative, and adversarial learning strategies to extract rich semantic information from medical images\cite{haghighi2022dira}.  More recent works aim at developing a universal SSL framework for medical images without tailoring pretext tasks. Xie et al. propose a universal pretraining method UniMiSS, leveraging transformer’s sequence-to-sequence modeling to pretrain on both 2D and 3D medical images \cite{xie2022unimiss}. As another universal framework, MIM also gains attention in medical image analysis. Chen et al. first explore using SimMIM and MAE for 3D medical image segmentation \cite{chen2023masked}. However, MIM still remains data-intensive and requires large datasets for scaling up model parameters \cite{xie2023data}. Furthermore, the success of MIM heavily relies on the mask strategies due to the spatial information redundancy in images \cite{he2022masked}. Intuitively, a universal medical MIM task should guide the model to focus on anatomically rich areas for reconstruction. Consequently, the model gains a deeper understanding of the anatomical information to generate masks for more effective learning.

\section{Method}
\subsection{Overview}
Figure 2 illustrates the overall pipeline of AnatoMask, which is composed of a student network and a teacher network sharing the same underlying architecture. Each module has an image encoder and an image decoder. Given an input CT volume at epoch $t$, we first generate a random mask with masking ratio $\gamma$ similar to SimMIM \cite{xie2022simmim}. Then, we feed the randomly masked input into the teacher network, whose encoder \( E_T \) abstracts the input into embeddings and its decoder \( D_T \) generates the reconstruction. We then compute the reconstruction loss \( L_{rec}^t \) on the masked regions by subtracting with input. Next, we sort  \( L_{rec}^t \) in descending order and use a masking dynamics function \(r_t\) to select top anatomical regions with high losses to form a binary mask \( M_{top}^t \). Then, we generate \( M_{\textit{final}}^t \) by randomly masking the remaining regions to maintain an overall masking ratio $\gamma$. Finally, we apply the newly generated mask to the input and train the student network on reconstruction loss. To ensure a consistent mask generation between student and teacher, we update the teacher network with an exponential moving average of the student's network weights:
\begin{align}
{\theta}_{T}^{t+1} &= m{\theta}_{T}^{t}+(1-m){\theta}_{S}^{t+1}, \label{eq:2}
\end{align}
where $m$ represents the weight decay, ${\theta}_{T}^{t}$ and ${\theta}_{T}^{t+1}$ represent the teacher’s parameters at time $t$ and $t+1$ respectively and ${\theta}_{S}^{t}$ represents the student’s parameter at time $t$.

\begin{figure}[tb]
  \centering
  \includegraphics[width=1\linewidth]{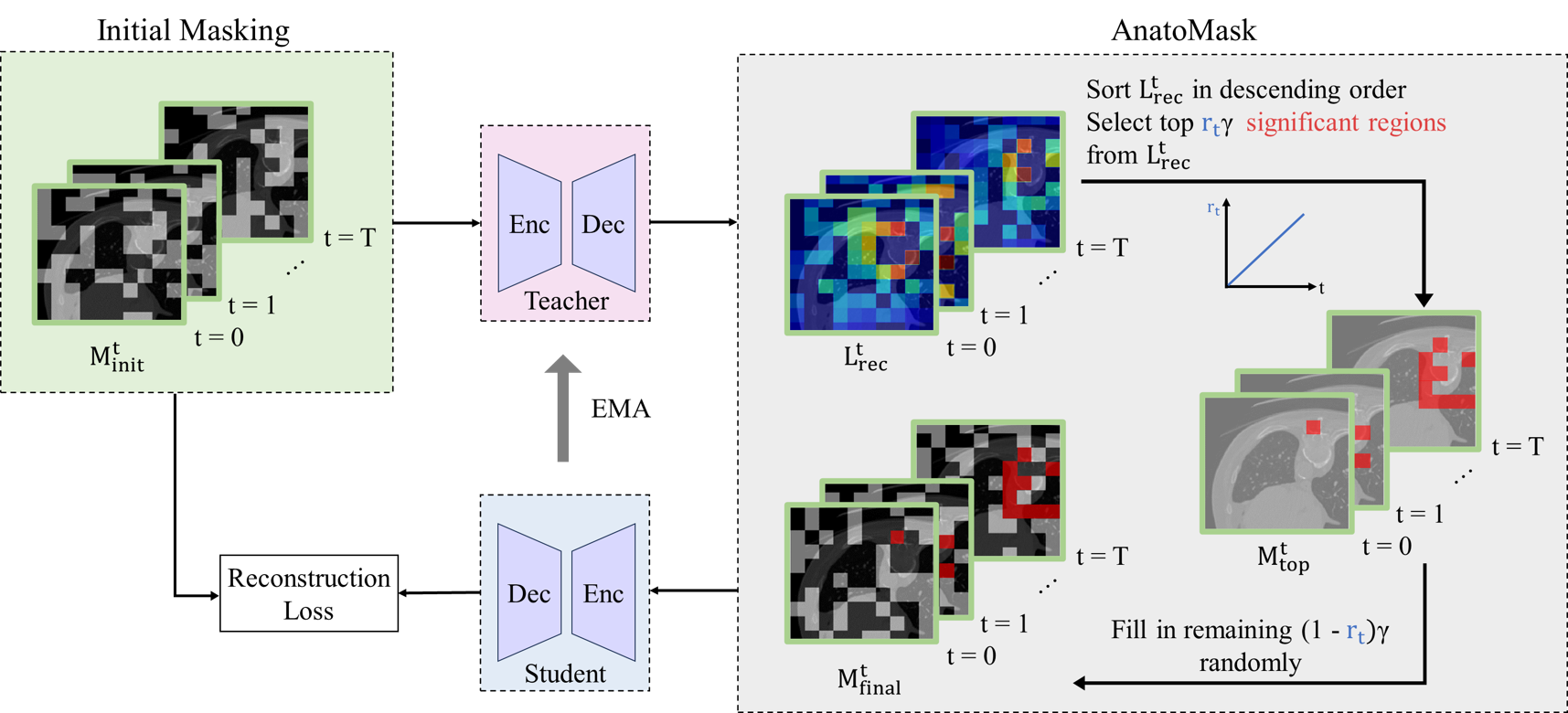}
  \caption{Overview of proposed AnatoMask. During SSL pretraining at epoch $t$, the teacher network receives randomly masked inputs and computes the patch-level reconstruction loss $L_{rec}^t$. The top $r_t$ regions with the highest reconstruction losses are selected to form a binary mask $M_{top}^t$. Then, the remaining $(1-r_t)\gamma$ areas are randomly filled with binary values to form our final mask $M_{\textit{final}}^t$. The student network is trained to reconstruct input masked by $M_{\textit{final}}^t$. }
  \label{fig:example}
\end{figure}

\subsection{Hierarchical Image Encoder-decoder}
Conventional ViT-based MIM backbones such as MAE or SimMIM utilize a simple decoder design with a few linear layers. However, Tian et al. propose that a hierarchical encoder-decoder design performed better when pretraining on CNNs \cite{tian2023designing}. To this end, we also adopt a multi-scale encoder-decoder backbone for our AnatoMask, which is well-suited to spatially aggregate feature maps. We find that this approach enables the progressive reconstruction of the masked image, effectively incorporating both local and global features. Given an input volume \textit{V} of  $H \times W \times D$, we first divide \textit{V} input patches of $\frac{H}{16} \times \frac{W}{16} \times \frac{D}{16}$ and randomly masked 60\%. Then, we utilize STU-Net to downsample the masked input into 4 scales: $\frac{H}{2} \times \frac{W}{2} \times \frac{D}{2}$, $\frac{H}{4} \times \frac{W}{4} \times \frac{D}{4}$, $\frac{H}{8} \times \frac{W}{8} \times \frac{D}{8}$, $\frac{H}{16} \times \frac{W}{16} \times \frac{D}{16}$. 
Similar to SparK \cite{tian2023designing}, we adopt the sparse submanifold convolution for our encoder to generate the multi-scale feature maps of the input. A hierarchical decoder is used to densify and upscale the feature maps for reconstruction. Concretely, let $M_{\textit{final}}^{t}$ denote the newly generated mask, the reconstruction loss can be formulated as:
\begin{align}
\mathcal{L}_{rec}^{t} &= \frac{1}{HWD}\sum(D_S(E_S(M_{\textit{final}}^{t}\odot V)) - (1-M_{\textit{final}}^{t})\odot V)^2, \label{eq:2}
\end{align}
where $E_S$ and $D_S$ represent the student network's encoder and decoder.

\begin{figure}[tb]
  \centering
  \includegraphics[height = 11cm]{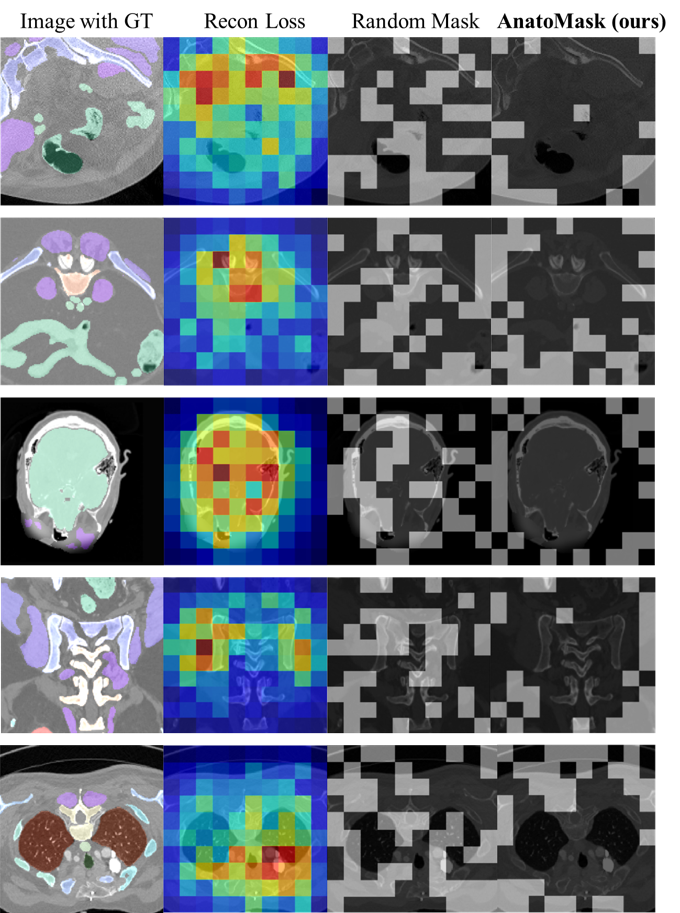}
  \caption{Visual comparison between segmentation ground truths and reconstruction losses. For each row, we show a). an image with organ ground truths, b). reconstruction losses by averaging over 2 masks, c). a random mask (60\%), and d). our AnatoMask obtained from b). Red means higher loss values while blue indicates lower ones. Red areas tend to overlap with organ regions. Transparent areas indicate unmasked regions.}
  \label{fig:AnatoMask_Fig3}
\end{figure}

\subsection{Reconstruction-guided Self-masking}
While recent studies in MIM have demonstrated remarkable successes in natural images \cite{xie2022simmim, he2022masked}, its effectiveness in medical imaging is worth further exploration due to the large domain gap between natural images and medical images. Conventional MIM takes a random masking approach, which ignores the anatomical region's heterogeneity by uniformly sampling all areas. Given the significant feature variations of human organs, a uniform random masking approach may not be an effective pretraining task for subsequent segmentation tasks. Thus, an effective MIM strategy should prioritize these complex anatomical regions to enhance diagnostic accuracy and efficacy. Intuitively, these complex regions are more difficult to reconstruct than background regions. We then take the regions with high reconstruction losses as the highly informative areas to guide the masking process. As demonstrated in \cref{fig:AnatoMask_Fig3}, 
we find that anatomical regions often demonstrate high reconstruction losses, indicative of rich semantic information for the network to learn. 

To identify such regions, a naive approach is to use a pretrained model as a teacher to extract high-loss areas from its predictions, which remains resource-intensive. Nevertheless, this frozen teacher cannot adaptively refine the representations of masked regions, which could hinder the student from learning more diverse features. Alternatively, we propose a self-distillation approach where the model first serves as a teacher to generate anatomically-significant masks and then acts as a student to perform self-supervised learning. In this way, we can train the model to self-mask anatomically informative regions, subsequently guiding the MIM toward more efficient pretraining. 

Given an initial mask, our AnatoMask teacher can generate an initial reconstruction loss $L_{rec}^t$, from which we can employ anatomical priors in its masking process to enhance pretraining. We perform an \textit{argsort} operation on $L_{rec}^t$ to rank the masked regions in \textit{descending} reconstruction losses. This ranking guides the model to focus on more challenging regions for reconstruction, which are more valuable for mining anatomical information. In the initial stages of training, high reconstruction loss may not always correlate with anatomical importance, as the model is still learning to reconstruct the image effectively. We propose a dynamic mask generation strategy, where the portion of anatomically significant regions in the masked regions gradually increases, providing some reasonable hints that guide the model to reconstruct anatomical regions step by step. By incrementally guiding the model, we ensure it learns effective representations, aligning more closely with anatomical significance as it advances in the training process. Concretely, we define a masking dynamics function based on current epoch $t$ (with total epoch $T$):
\begin{align}
r_t &= r_0 + \frac{t}{T}(r_T - r_0), \label{eq:2}
\end{align}
where $r_t$ represents anatomical-significance ratio of $M_{final}^{t}$. To retain the original masking ratio $\gamma$, the remaining $(1-r_t)$ of the masked regions in $M_{final}^{t}$ are "filled" by a newly generated random mask. $r_T$, and $r_0$ are two hyperparameters which we empirically set to 0.5 and 0 respectively. Given input volume \textit{V}, masking ratio $\gamma$, and $M_{initial}^t$ being the initial mask fed into the teacher model, we select $r_t \cdot \gamma V$ voxels with the highest $L_{rec}^t$ from $M_{initial}^t$ and then randomly generate $(1-r_t) \cdot \gamma V$ voxels from $V-r_t \cdot \gamma V$ to form the final mask $M_{final}^{t}$

\begin{algorithm}
\caption{Algorithm for AnatoMask \( M_{\text{final}}^t \)}
\begin{algorithmic} % The number tells where the line numbering should start
\State \textbf{input}: \( G = (g_{1}, g_{2},\ldots, g_{n}) \): All regions in the image.
\Statex \(\phantom{\textbf{input}: }\)\( L_{rec} = (l_{g_{1}}, l_{g_{2}},\ldots, l_{g_{n}}) \): Reconstruction losses from the initial mask.
\Statex \(\phantom{\textbf{input}: }\) \( \gamma \): Masking ratio.
\Statex \(\phantom{\textbf{input}: }\) \( r_{t} \): Anatomical-significance ratio.
\State \textbf{output}:  \( M_{\text{final}}^t \): Mask with anatomically-significant regions.

% \State \textbf{Sort} $G$ by $L_{rec}$ in descending order to get $G'$ 
% \State $j = argsort(L_{rec})$

% \State $K \gets {\rm argsort}(L_{\rm rec})$ \Comment{Sort losses}

\State $i \gets 1$
\State \textbf{for} $k$ in ${\rm argsort}(L_{\rm rec}):$
\Comment{Sort regions based on loss}
\State \quad $g'_i \gets g_k $
\State \quad $i \gets i + 1$
\State\textbf{endfor}
% for $i \in \{1,\dots,n\}$ 

% \State \textbf{Sort} \( L_{rec}\) in descending order to obtain \( L_{rec}^{'} = (l_{1}^{'}, l_{2}^{'},\ldots, l_{n}^{'}) \), where \( (l_{1}^{'} \geq l_{2}^{'} \ldots \geq l_{n}^{'}) \);
% \State \textbf{Calculate} the index \(i\) for the cutoff in \(L_{rec}'\) using \(i = \lceil \gamma^t \times n \rceil\);
\State \(p \gets \lceil r_t \times n \rceil\) 
\Comment{Compute selection threshold $p$}
\State \(M_{\text{top}} \gets \{\} \)  
% \Comment{Initialize} 
\State \textbf{for }\( i \) \text{ in } \(\{1,\dots,p\}\): \Comment{Select top $p$ regions}
\State \quad $M_{\text{top}}\gets M_{\text{top}} \cup g'_i$
% \State \quad \textbf{Add} \(G'[i]\) \text{to} \( M_{\text{top}} \) 
\State\textbf{endfor}
\State \( n_{\text{random}} \gets \lceil (\gamma-r_t) \times n \rceil \)  
\Comment{Compute how many regions to randomly select.}
% \State \textbf{Randomly select} \( n_{\text{random}} \) regions from \( G\setminus M_{\text{top }} \) to form \( M_{\text{random}} \).
\State \(M_{\text{random}} \gets \{\} \) 
\State \textbf{for} $i$ in $\{1,\dots,n_{\rm random}\}:$ 
\Comment{Randomly sample $n_{random}$  regions}
\State  \quad $M_{\text{random}} \gets  M_{\text{random}} \cup \rm{sample}$ $ (G\setminus (M_{\text{top}} \cup M_{random} ))$
\State\textbf{endfor}
\State  $ M_{\text{final}}^t \gets M_{\text{top}}  \cup M_{\text{random}}$
\State \textbf{Return} \( M_{\text{final}}^t \)

\end{algorithmic}
\end{algorithm}

\section{Result}
\subsection{Dataset and Evaluation}
We conduct self-supervised pretraining for all methods on TotalSegmentator \cite{wasserthal2023totalsegmentator} dataset containing 1204 CT images with 104 anatomical structures (27 organs, 59 bones, 10 muscles, and 8 vessels). For preprocessing, all the images are resampled to $1.5 \times 1.5 \times 1.5$ $mm^3$ and z-score normalization is used to rescale image intensity to zero mean and unit variance. Image patch size is set to $112 \times 112 \times 128$. We first use the following ratio to randomly split the dataset: 70\% train, 10\% validation, and 20\% test. The training set is used for SSL pretraining. We then finetune the pretrained networks on the training set and use the validation set for early stops. 

We further conduct extensive evaluations by finetuning on 3 public datasets to test the transferability of our models. We select FLARE22 (50 CT cases of 13 organs) \cite{ma2022fast}, AMOS22 (240 CT cases and 120 MR cases with 15 organ annotations) \cite{ji2022amos} and AutoPETII (398 PET/CT cases with lesion annotations) \cite{gatidis2022whole}. For each dataset, we utilize 70\% train, 10\% validation, and 20\% test for evaluation. 

For evaluation metrics, we calculated the Dice Coefficient (DSC) and Normalized Surface Dice (NSD) based on segmentation results. 

\subsection{Implementation Details}
All experiments in this study are conducted based on Python 3.9, Ubuntu 22.04.3, Pytorch 2.0.1, and nnU-Net 2.2 \cite{isensee2021nnu}. For finetuning on TotalSegmentator, we utilize default data pre-processing and data augmentations in nnU-Net. For SSL pretraining, we follow the same data pre-processing as in TotalSegmentator but do not use additional data augmentations other than random cropping and random flipping. For finetuning on FLARE22, AMOS22, and AutoPETII, we use the nnUNet’s default data pre-processing and data augmentation pipelines for each dataset. However, we use the same network configuration as in TotalSegmentator and in SSL pretraining. A more detailed description of hyperparameters is shown in the supplementary material. Models are trained on NVIDIA Tesla A100 cards with 80 GB VRAM. 

\subsection{Ablation Study}
In this section, we investigate the decoder design, different masking strategies, and masking ratios of AnatoMask. We use STU-Net-B as the backbone with 1000 epochs for pretraining and finetuning on TotalSegmentator.

\textbf{Decoder design.} First, we verify the effectiveness of using a hierarchical decoder for AnatoMask. We compare the U-Net style architecture with a simple decoder architecture, which only uses two convolution blocks and does not receive hierarchical feature maps from the encoder. As shown in \cref{tab:ablation1}, a U-Net style encoder-decoder performed better than a simple decoder for MIM. Nevertheless, our AnatoMask still improves finetuning performance on a simple decoder design compared to random masking. 

\begin{table}[tb]
  \caption{Ablation on the encoder-decoder design of AnatoMask.
  }
  \label{tab:ablation1}
  \centering
  \begin{tabular}{@{\hspace{10pt}}l@{\hspace{10pt}}l@{\hspace{10pt}}l@{}}
    \toprule
    Architecture & Masking & DSC (\%)\\
    \midrule
    Simple Decoder & Random & 82.9\\
    Simple Decoder & AnatoMask & 83.4\\
    Hierarchical Decoder & Random & 83.6\\
    \textbf{Hierarchical Decoder} & \textbf{AnatoMask} & \textbf{84.3}\\
  \bottomrule
  \end{tabular}
\end{table}

\textbf{Effectiveness of Self-distillation.} Then, we verify the importance of using self-distillation to maintain consistent mask generations between teacher and student. In our proposed approach, the teacher network is updated with student's EMA weights at the end of each iteration. Without this, the teacher is simply a randomly initialized model without explicit knowledge on where to mask. As shown in \cref{tab:ablation2} rows 2 and 3, self-distillation ensures that the teacher consistently generates meaningful MIM tasks for the student to solve. 

\textbf{Incorporating anatomically-significant regions.} We study the effectiveness of reconstruction guidance to incorporate anatomically-significant regions. To obtain regions of high significance for masking, we apply \textit{argsort} operations on the reconstruction losses in \textit{descending order} to obtain the top $r_t$ regions. Conversely, if we sort the reconstruction losses in \textit{ascending} order, we then obtain regions with low significance, possibly making the final mask easier than a random mask to restore. As shown in \cref{tab:ablation2} rows 3 and 4, we find that models did not benefit from learning low-significance regions, demonstrating the importance of incorporating anatomically rich regions. 

\textbf{Effectiveness of masking dynamics function.} Additionally, we study the importance of our proposed masking dynamics function which controls the MIM difficulties over time. Specifically, the relationship between $r_0$ and $r_T$ determines whether the pretraining difficulties increase or decrease with time. When $r_0>r_T$, more anatomically complex components are included in the final mask, and fewer components are generated from random masking. This leads to an easy-to-hard objective which is more intuitive. Conversely, if $r_0<r_T$, we obtain a hard-to-easy MIM objective. As shown in \cref{tab:ablation2}, we observe that an easy-to-hard masking strategy leads to better finetuning performance. This aligns with our hypothesis that starting with a low masking difficulty can offer valuable cues to the model and prevent training collapse. 

\begin{table}[tb]
  \caption{Ablation study on the masking strategies of AnatoMask. We study the importance of our easy-to-hard masking dynamics, and self-distillation scheme and select the highest or lowest anatomically significant regions.}
  \label{tab:ablation2}
  \centering
    \begin{tabular}{@{\hspace{5pt}}l@{\hspace{10pt}}l@{\hspace{10pt}}l@{\hspace{10pt}}l@{\hspace{10pt}}l@{}}
    \toprule
    Self-distillation & Anatomical Significance & Masking Dynamics & DSC (\%)\\
    \midrule
    $\times$  & - & - & 83.6\\ 
    $\times$  & High & easy-to-hard & 83.7\\
   $\pmb{\checkmark}$ & \textbf{High} & \textbf{easy-to-hard} & \textbf{84.3}\\
    $\checkmark$  & Low & easy-to-hard & 83.6\\
  $\checkmark$ & High & hard-to-easy & 84.0\\
  \bottomrule
  \end{tabular}
\end{table}

\textbf{Masking ratio $\gamma$.} As shown in \cref{tab:ablation3}, we find that a lower masking ratio outperformed a higher masking ratio, different from traditional MIM methods. We hypothesize that a lower masking ratio leads to better preservation of the original anatomical information, resulting in more effective pretaining. 

\begin{table}[tb]
  \caption{Ablation study on masking ratio $\gamma$.}
  \label{tab:ablation3}
  \centering
  \begin{tabular}{@{\hspace{5pt}}l@{\hspace{10pt}}l@{}}
    \toprule
    Masking ratio $\gamma$ & DSC (\%)\\
    \midrule
    \textbf{0.6}  & \textbf{84.3}\\
    0.8  & 84.1\\
    0.9  & 84.2\\
  \bottomrule
  \end{tabular}
\end{table}

\begin{table}[tb]
  \caption{Comparison of AnatoMask and existing SSL methods on TotalSegmentator. Data is shown as patient-wise mean ± SD. \textbf{Bold} means best method}
  \label{tab:comparetotal}
  \centering
    \begin{tabular}{@{\hspace{5pt}}l@{\hspace{10pt}}l@{\hspace{10pt}}l@{\hspace{10pt}}l@{\hspace{10pt}}l@{}}
    \toprule
    Methods & Backbone & DSC (\%) & NSD (\%)\\
    \midrule
    Tang et al. \cite{tang2022self}  & SwinUNETR & $79.2\pm11.2$ & $75.3\pm12.8$\\
    SelfMedMAE \cite{zhou2023self} & UNETR & $78.8\pm12.5$ & $74.0\pm15.0$\\
    UniMiSS \cite{xie2022unimiss}& MiTnet & $82.2\pm10.4$ & $86.4\pm13.0$\\
    MoCov2 \cite{chen2020improved}& STU-Net-B & $81.8\pm10.6$ & $85.6\pm12.3$\\
    BYOL \cite{grill2020bootstrap}& STU-Net-B & $78.8\pm11.1$ & $82.8\pm12.7$\\
    SparK \cite{tian2023designing}& STU-Net-B & $83.6\pm10.5$ & $86.7\pm12.8$\\
    \midrule
    \textbf{AnatoMask} & STU-Net-B & $\boldsymbol{84.3\pm9.9}$ & $\boldsymbol{87.8\pm11.8}$\\
    \textbf{AnatoMask} & STU-Net-L & $\boldsymbol{87.8\pm9.8}$ & $\boldsymbol{90.3\pm10.7}$\\
    \textbf{AnatoMask} & STU-Net-H & $\boldsymbol{89.0\pm10.3}$ & $\boldsymbol{91.9\pm10.0}$\\
  \bottomrule
  \end{tabular}
\end{table}

\subsection{Comparison with previous SSL methods}

\begin{figure}[tb]
  \centering
  \includegraphics[width=1\linewidth]{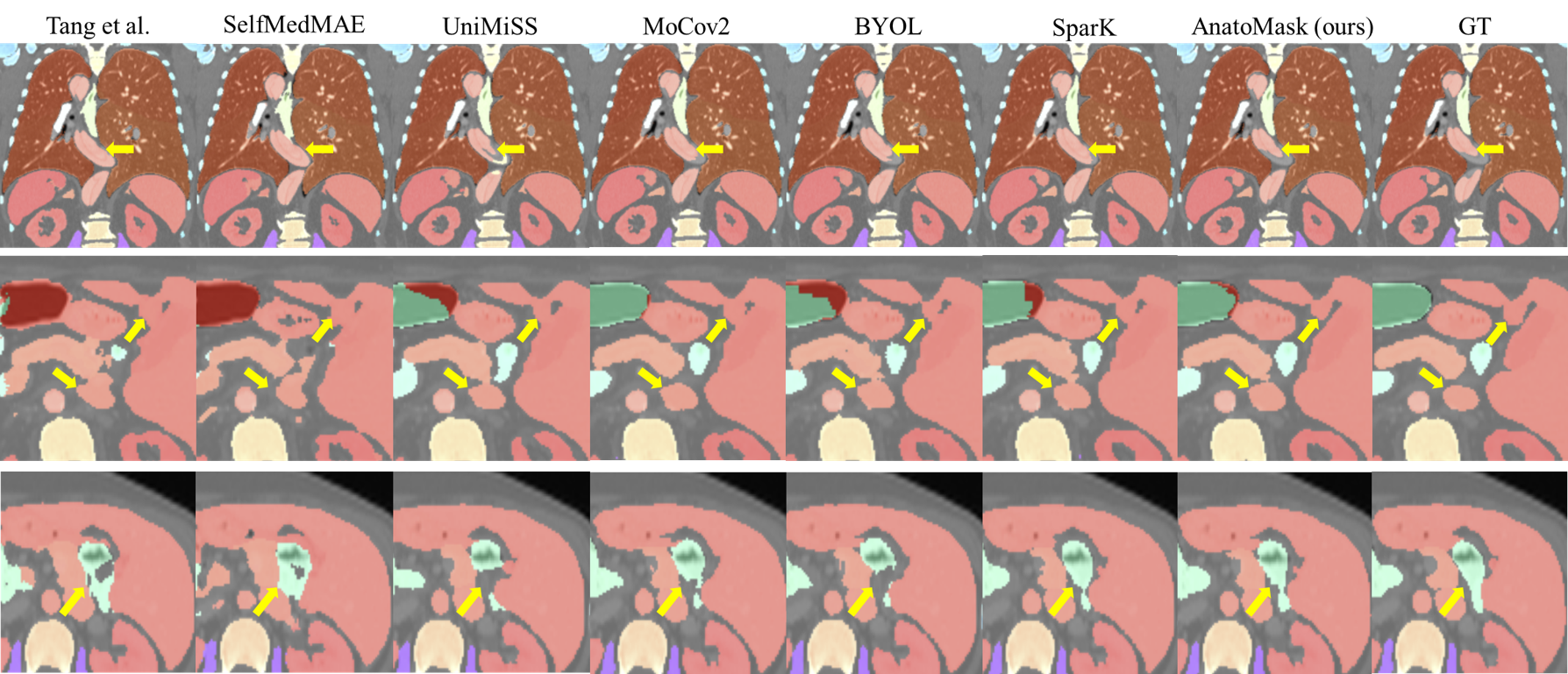}
  \caption{Visualization of multi-organ CT segmentation results on TotalSegmentator. Yellow arrows indicate improvements in segmentation.}
  \label{fig:totalseg}
\end{figure}
\begin{figure}[tb]
  \centering
  \includegraphics[width=1\linewidth]{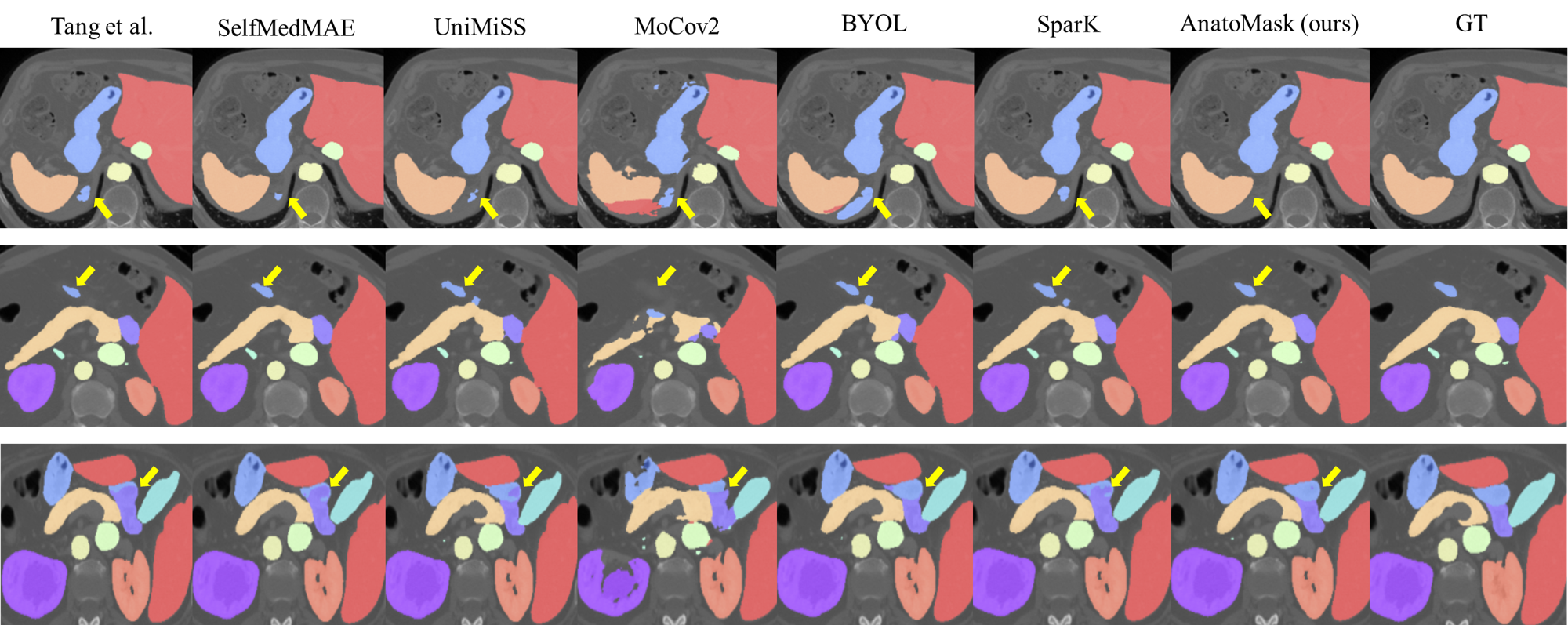}
  \caption{Visualization of multi-organ CT segmentation results on FLARE22. Yellow arrows indicate improvements in segmentation.}
  \label{fig:flare22}
\end{figure}

\begin{figure}[tb]
  \centering
  \includegraphics[width=1\linewidth]{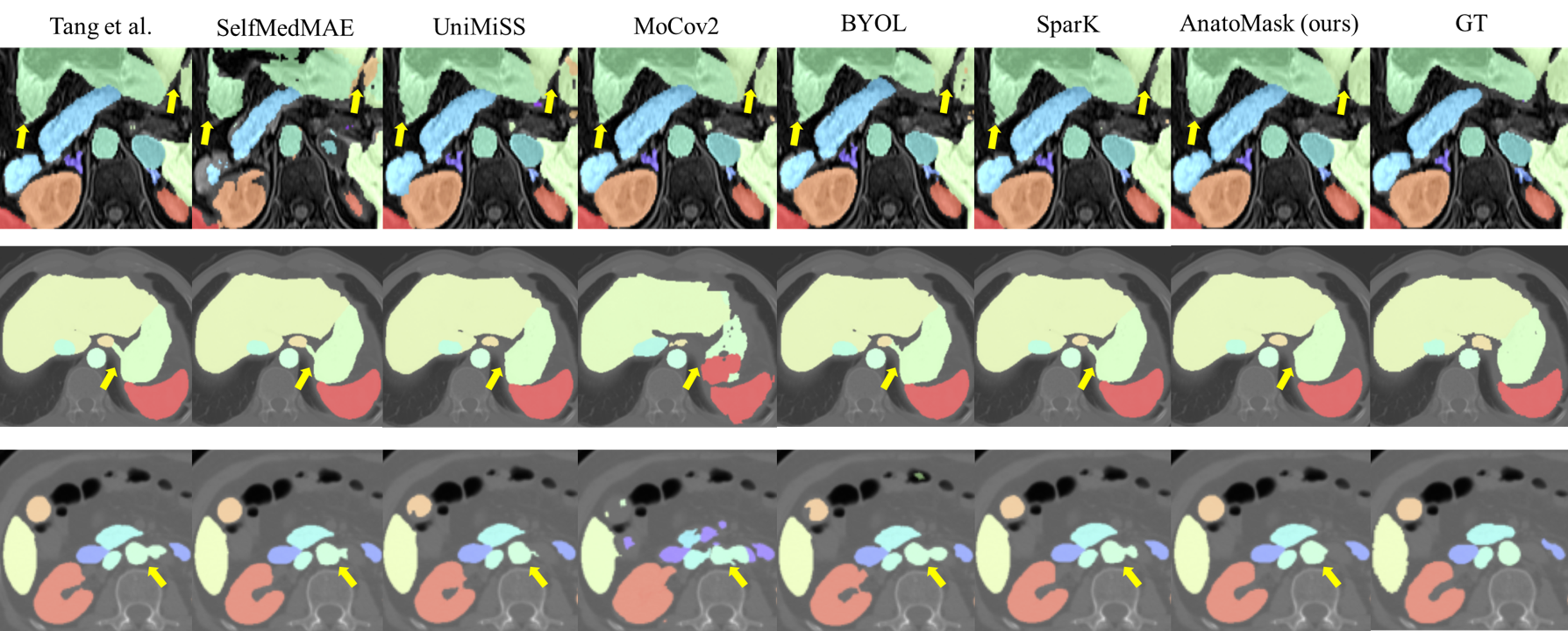}
  \caption{Visualization of cross-modality organ segmentation on MRI (first row) and CT (second and third rows) from AMOS22. Yellow arrows indicate improvements in segmentation.}
  \label{fig:amos22}
\end{figure}

\begin{figure}[tb]
  \centering
  \includegraphics[width=1\linewidth]{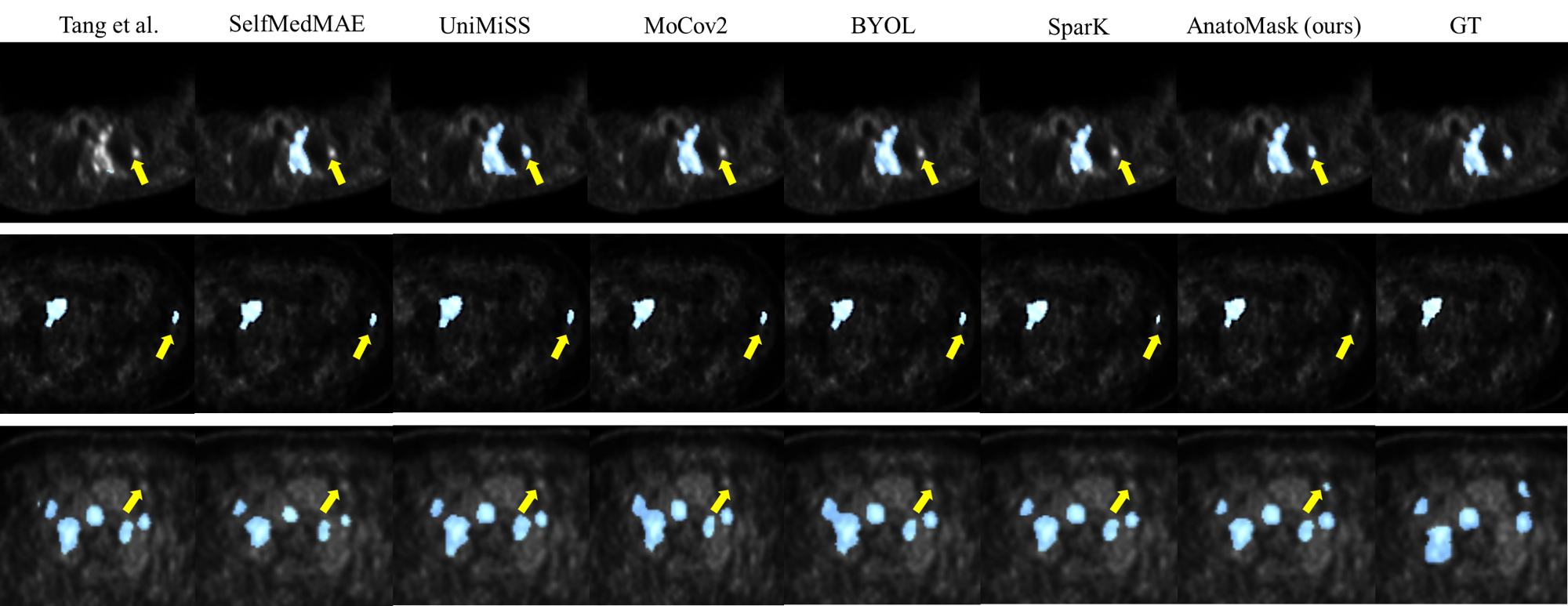}
  \caption{Visualization of PET lesion segmentation results on AutoPETII. Yellow arrows indicate improvements in segmentation.}
  \label{fig:autopet}
\end{figure}

We compare the finetuning performances of our method against previous SOTA SSL methods: 1). Tang et al. propose a combination of pretext tasks to self-supervise a transformer model SwinUNETR \cite{tang2022self}; 2). Zhou et al. propose SelfMedMAE, a masked autoencoder method to pretrain transformer models \cite{zhou2023self}; 3). Xie et al. propose UniMiSS to extract 2D information from 3D volumes using self-distillation pretraining \cite{xie2022unimiss}; 4). He et al. propose MoCov2, a momentum contrastive method that overcomes the limitations of large batch size \cite{chen2020improved}; 5). Grill et al. propose BYOL, a contrastive method with dual-network to learn bootstrapped representations \cite{grill2020bootstrap}; 6). Spark, a U-Net design for BERT-style pretraining of CNNs \cite{tian2023designing}. For network backbones, since certain methods are built on specific transformer networks, we maintain their backbones for fair comparison. For MoCov2 and BYOL, we use STU-Net-B as the backbone. For all methods, we fix pretraining epochs to 1000. The input patch size for SwinUNETR, UNETR, and MiTnet is set to (128, 128, 128) and the batch size is set to 4. 
As shown in \cref{tab:comparetotal}, AnatoMask with STU-Net-B as backbone significantly outperforms all previous SSL methods. With 1000 epochs for pretraining, AnatoMask is able to achieve 84.3\% DSC and 87.8\% SDC, surpassing the best method (UniMiSS) by 2.1\% and 1.4\% respectively. Notably, UniMiSS extracts both 2D information and 3D information from medical images, whereas our AnatoMask focuses only on 3D information. This further demonstrates AnatoMask’s superior efficiency and effectiveness in leveraging 3D data for segmentation tasks. We visualize segmentation results on TotalSegmentator CT images in \cref{fig:totalseg}. Our Anatomask demonstrates promising segmentation results compared with other SSL methods. To demonstrate the scalability of AnatoMask, we further test it on large segmentation backbones such as STU-Net-L and STU-Net-H (with 440 million and 1.5 billion parameters respectively) which improve performance to 87.8\% DSC and 89.0\% DSC. Our results indicate that AnatoMask is also beneficial for network parameter scaling, paving the way for self-supervising large foundation models for medical image segmentation.

{
\begin{table}[tb]
  \caption{Comparison of AnatoMask and existing SSL methods when transferring to FLARE22 and AMOS22. Data is shown as patient-wise mean ± SD. \textbf{Bold} means best method.}
  \label{tab:comparetransfer1}
  \centering
  {
  \begin{tabular}{@{\hspace{1pt}}l@{\hspace{5pt}}l@{\hspace{5pt}}c@{\hspace{5pt}}c@{\hspace{5pt}}c@{\hspace{5pt}}c@{\hspace{5pt}}c@{\hspace{5pt}}}
  \toprule
  Methods & Backbone & \multicolumn{2}{c}{FLARE22} & \multicolumn{2}{c}{AMOS22}\\
      \cmidrule(lr){3-4} \cmidrule(lr){5-6} & & DSC & NSD & DSC & NSD \\
  \midrule
  Tang et al. \cite{tang2022self} & SwinUNETR & $91.2\pm2.1$ & $94.1\pm2.6$ & $85.4\pm5.8$ & $91.3\pm6.1$\\
  SelfMedMAE \cite{zhou2023self} & UNETR & $91.0\pm3.4$ & $93.7\pm3.2$ & $76.1\pm7.7$ & $86.7\pm6.0$\\
  UniMiSS \cite{xie2022unimiss} & MiTnet & $91.7\pm2.2$ & $95.2\pm2.2$ & $87.7\pm5.6$ & $93.7\pm5.5$\\
  MoCov2 \cite{chen2020improved} & STU-Net-B & $93.6\pm1.3$ & $97.0\pm1.6$ & $88.3\pm5.3$ & $94.1\pm4.9$\\
  BYOL \cite{grill2020bootstrap} & STU-Net-B & $93.7\pm1.2$ & $97.0\pm1.4$ & $88.3\pm5.4$ & $94.1\pm4.7$\\
  SparK \cite{tian2023designing}& STU-Net-B & $93.8\pm1.2$ & $97.1\pm1.2$ & $88.5\pm5.3$ & $94.2\pm4.8$\\
  \midrule %
  \textbf{AnatoMask} & STU-Net-B & $\boldsymbol{94.0\pm1.3}$ & $\boldsymbol{97.3\pm1.5}$ & $\boldsymbol{89.1\pm5.0}$ & $\boldsymbol{94.8\pm4.5}$\\
  \textbf{AnatoMask} & STU-Net-L & $\boldsymbol{94.2\pm1.1}$ & $\boldsymbol{97.3\pm1.3}$ & $\boldsymbol{89.9\pm5.4}$ & $\boldsymbol{95.6\pm4.7}$\\
  \textbf{AnatoMask} & STU-Net-H & $\boldsymbol{94.3\pm1.5}$ & $\boldsymbol{97.3\pm1.5}$ & $\boldsymbol{89.9\pm5.2}$ & $\boldsymbol{95.7\pm5.0}$\\
  \bottomrule
  \end{tabular}
  }
\end{table}
}

Finally, we conduct transfer learning on three additional datasets by finetuning each model trained on TotalSegmentator. To demonstrate the transferability of our method, we choose three segmentation tasks: 1). multi-organ segmentation on CT (FLARE22); 2). cross-modality segmentation on MRI and CT (AMOS22 Task 2); 3). tumor segmentation on PET/CT (AutoPETII). We initialized all models with weights trained from TotalSegmentator and finetuned for 1000 epochs. As shown in \cref{tab:comparetransfer1} and \cref{tab:comparetransfer2}, AnatoMask with STU-Net-B as backbone consistently outperforms comparable methods. We also visualize 
multi-organ segmentation results on CT in \cref{fig:flare22}, cross-modality segmentation results (CT and MRI) in \cref{fig:amos22}, and PET lesion segmentation results in \cref{fig:autopet}. These results further demonstrate that our SSL strategy possesses promising transferability to other modalities, such as MRI and PET. When using larger backbones STU-Net-L and STU-Net-H, AnatoMask further improves segmentation performance on each task, demonstrating the scalability of our method. 

{
\begin{table}[tb]
\caption{Comparison of AnatoMask and existing SSL methods when transferring to AutoPETII. Data is shown as patient-wise mean ± SD. \textbf{Bold} means best method.}
  \label{tab:comparetransfer2}
  \centering
  {
  \begin{tabular}{@{\hspace{1pt}}l@{\hspace{5pt}}l@{\hspace{5pt}}c@{\hspace{5pt}}c@{\hspace{5pt}}}
  \toprule
  Methods & Backbone & \multicolumn{2}{c}{AutoPETII} \\
  \cmidrule(lr){3-4}
  & & DSC & NSD \\
  \midrule
  Tang et al. \cite{tang2022self} & SwinUNETR & $72.3\pm23.7$ & $80.5\pm22.8$\\
  SelfMedMAE \cite{zhou2023self}& UNETR & $67.0\pm23.9$ & $74.8\pm24.3$\\
  UniMiSS \cite{xie2022unimiss}& MiTnet & $64.0\pm23.7$ & $75.0\pm23.0$\\
  MoCov2 \cite{chen2020improved}& STU-Net-B & $72.9\pm21.3$ & $81.2\pm21.6$\\
  BYOL \cite{grill2020bootstrap}& STU-Net-B & $74.1\pm21.1$ & $82.2\pm21.7$\\
  SparK \cite{tian2023designing}& STU-Net-B & $75.3\pm21.1$ & $84.5\pm21.4$\\
  \midrule % This adds a line after SparK
  \textbf{AnatoMask} & STU-Net-B & $\boldsymbol{75.8\pm19.7}$ & $\boldsymbol{85.3\pm19.7}$\\
  \textbf{AnatoMask} & STU-Net-L & $\boldsymbol{77.5\pm19.0}$ & $\boldsymbol{85.9\pm18.7}$\\
  \textbf{AnatoMask} & STU-Net-H & $\boldsymbol{78.1\pm19.7}$ & $\boldsymbol{87.0\pm19.8}$\\
  \bottomrule
  \end{tabular}
  }
\end{table}
}

\section{Conclusion and Discussion}
In this paper, we propose AnatoMask, a novel SSL approach to augment the existing MIM workflow for enhancing medical image segmentation. Compared to conventional random masking techniques, AnatoMask aims at mining anatomical information through an iterative masking process, guided by the reconstruction loss which signifies the anatomically significant regions. We find that AnatoMask not only augments the quality of representations learned but also boosts data efficiency in the subsequent finetuning phase. Our experiments on 4 public datasets demonstrate that AnatoMask consistently improves the segmentation performance of different organs and imaging modalities. Our results also show strong scalability with larger segmentation backbones. 

However, our study still has some limitations. We only conduct SSL pretraining on CT images. Given the availability of a vast array of unlabeled MR images, we plan to expand our SSL approach to this modality. Also, our evaluation focuses on organ segmentation in this paper. However, given the encouraging transferability observed in PET lesion segmentation, our future study will explore the development of foundational models for more comprehensive tumor segmentation tasks in CT and MR images. 

\clearpage  % TODO REVIEW/FINAL: This \clearpage needs to be removed from both review and camera-ready versions.

\section*{Acknowledgements}
This research is supported in part by the National Institutes of Health under Award Number R01CA272991, R56EB033332 and R01EB032680.

\bibliographystyle{splncs04}
\bibliography{main}
\end{document}